\documentclass{article}

\usepackage{arxiv}

\usepackage[utf8]{inputenc} 
\usepackage[T1]{fontenc}    
\usepackage{hyperref}       
\usepackage{url}            
\usepackage{booktabs}       
\usepackage{amsfonts}       
\usepackage{nicefrac}       
\usepackage{microtype}      
\usepackage{lipsum}         
\usepackage{graphicx}
\usepackage{doi}
\usepackage{mathtools}
\usepackage[table,xcdraw]{xcolor}
\usepackage{multirow}
\usepackage{amsmath}
\usepackage{cleveref}       
\usepackage{bbm}
\usepackage{algorithm}
\usepackage{algorithmic}

\title{Decision Mamba Architectures}

\date{}

\author{ \hspace{1mm}André Correia \\
	Universidade da Beira Interior and NOVA LINCS \\
	Covilhã, Portugal \\
	\texttt{andre.correia@ubi.pt} \\
	\And
	\hspace{1mm}Luís A. Alexandre \\
	Universidade da Beira Interior and NOVA LINCS \\
	Covilhã, Portugal \\
	\texttt{luis.alexandre@ubi.pt} \\
}

\renewcommand{\shorttitle}{Decision Mamba Architectures}

\hypersetup{
pdftitle={Decision Mamba Architectures},
pdfsubject={q-bio.NC, q-bio.QM},
pdfauthor={André Correia, Luís A. Alexandre},
pdfkeywords={Demonstration, Reinforcement, Learning, Sequence, Modelling},
}

\begin{document}
\maketitle

\begin{abstract}
Recent advancements in imitation learning have been largely fueled by the integration of sequence models, which provide a structured flow of information to effectively mimic task behaviours. Currently, Decision Transformer (DT) and subsequently, the Hierarchical Decision Transformer (HDT), presented Transformer-based approaches to learn task policies. Recently, the Mamba architecture has shown to outperform Transformers across various task domains. In this work, we introduce two novel methods, Decision Mamba (DM) and Hierarchical Decision Mamba (HDM), aimed at enhancing the performance of the Transformer models. Through extensive experimentation across diverse environments such as OpenAI Gym and D4RL, leveraging varying demonstration data sets, we demonstrate the superiority of Mamba models over their Transformer counterparts in a majority of tasks. Results show that DM outperforms other methods in most settings. The code can be found at https://github.com/meowatthemoon/DecisionMamba.
\end{abstract}

\keywords{Machine Learning, Demonstration Learning, Sequence Modelling, Imitation Learning}

\section{Introduction}
\label{introduction}

Reinforcement learning (RL) \cite{barto} has showcased remarkable prowess across a broad spectrum of robotic tasks, ranging from object manipulation like pushing and grasping to complex navigational challenges such as path finding and locomotion \cite{stabilizing,conservative,ris}.
However, the inherent trial-and-error nature of online learning, crucial for estimating policies in large state and action spaces, poses a substantial cost in real-world applications. 
Furthermore, the process of learning a policy through online RL requires the crafting of a tailored reward function specifically designed to guide exploration. This requirement adds another layer of complexity to the problem, due to the difficulty of devising a function that comprehensively covers the entire state and action space.

Offline RL emerges as a promising solution to unlock RL's full potential by circumventing both the need for active engagement with an online environment and the creation of a reward function. By leveraging pre-existing demonstration data sets, offline RL offers a pathway to effective and generalizable policy learning \cite{levine}. 
Specifically, Behavioural Cloning (BC) formulates the problem of learning the task as a supervised learning problem to imitate the policies represented in data set.

Transformer models have sparked a paradigm shift across various machine learning domains and have initially been applied to RL in the form of decision Transformers (DT) \cite{dt}. 
DTs cast RL problems into sequence modelling, empowering agents to assimilate a series of past interactions rather than a single observation, allowing them to make more informed decisions.
However, studies \cite{hdt} have shown that DT relies on the reward sequence for guidance, which requires the specification of the value of desired accumulated rewards that is task specific and non-trivial. Moreover, the reliance on deterministic rewards causes the method to fail in stochastic environments \cite{td}.

\begin{figure}[!tb]
\centering
\includegraphics[width=0.7\linewidth]{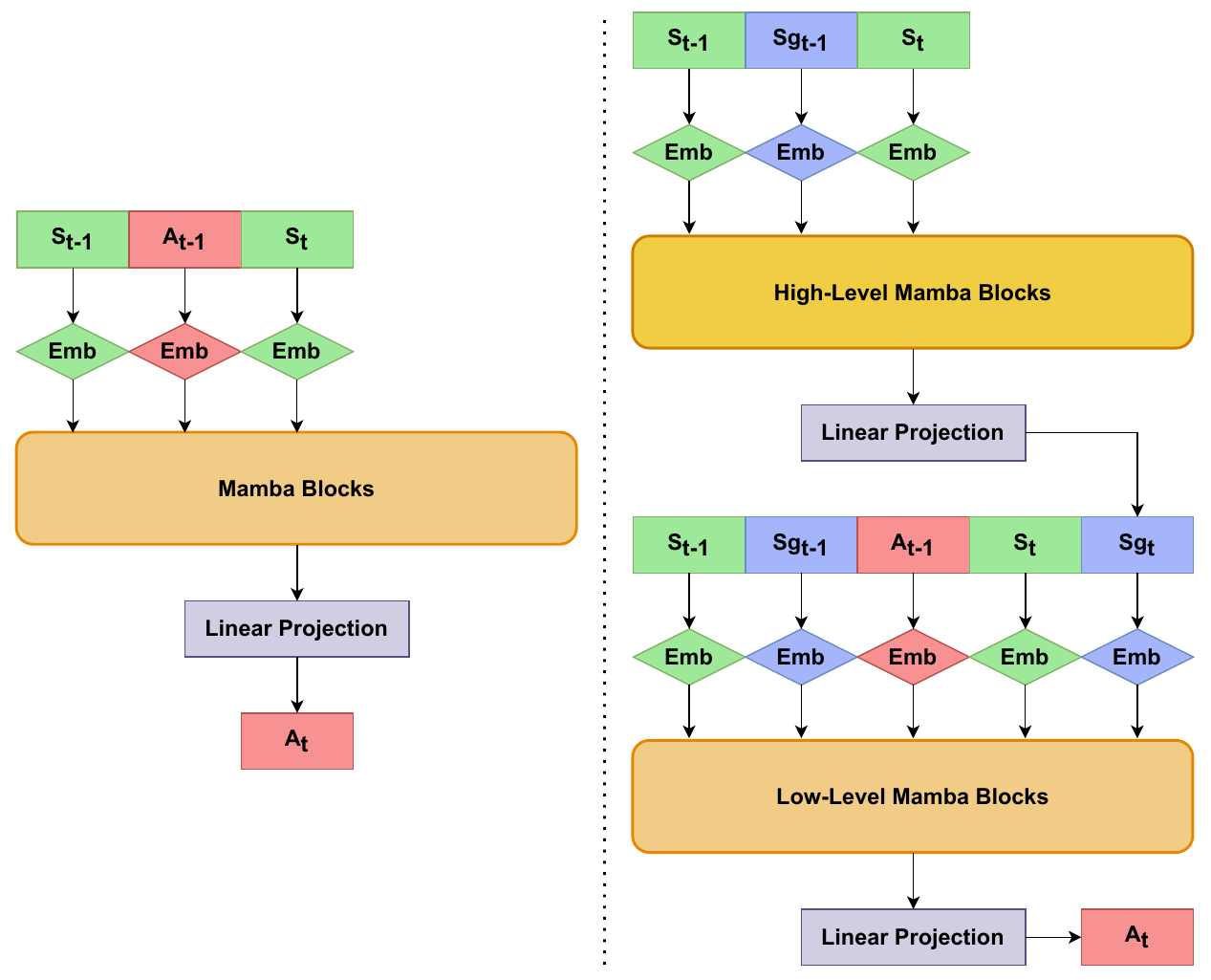}
\caption{The DM architecture on the left and the HDM architecture on the right side. The DM is conditioned on the sequence of past states and actions to predict the correct action. The HDM is composed of two modules. The high-level mechanism guides the low-level controller through the task by selecting sub-goal states, based on the history of sub-goals and states. The low-level controller is conditioned on the history of past states, sub-goals, and actions to select the appropriate action.}
\label{fig:hdm}
\end{figure}

Although recent methods have tackled these issues by substituting the reward sequence \cite{hdt,td}, they require extra models for this purpose. Specifically, Hierarchical Decision Transformer replaces the reward sequence with a sub-goal sequence proposed by a higher level Transformer.

Recently, structured state space sequence models (SSMs) \cite{ssms}, have garnered attention for their linear scalability with sequence length, outperforming Transformers in a wide range of domains. We identify that the evolutionary parameter of SSMs provides the guidance signal that the sequence of rewards offers DT. For this reason we propose to replace the Transformer architecture of both DT and HDT resulting in the Decision Mamba (DM) and Hierarchical Decision Mamba (HDM) models. Specifically for DM, we show that it can provide the sequence modelling advantages of the original DT without requiring the sequence of rewards which requires task knowledge and user intervention as well as causing it to fail in stochastic environments.

We conduct experiments on seven tasks from D4RL benchmark. Our results show that the performance of the DM is no longer dependent on the sequence of rewards, requiring no user intervention or task knowledge. Unlike the other methods, the DM does not require access to the reward function. Both DM and HDM outperform their Transformer state-of-the-art predecessors.

\textbf{Summary of Contributions:}
\begin{enumerate}
    \item We present the HDM and DM, the Mamba successors of HDT and DT, respectively. The models are smaller, faster and more accurate than their Transformer predecessors.

    \item We show that DM does not require the sequence of rewards to perform, unlike its predecessor DT. This removes the need for a reward function, task knowledge to craft an adequate desired reward value and user intervention.

    \item We evaluate the HDM and DM on seven environments and data sets from the D4RL \cite{d4rl} benchmark. Our methods outperform the original Transformer baselines, proving the viability of the Mamba architecture for imitation learning. Recent advancements in imitation learning have been largely fueled by the integration of sequence models, which provide a structured flow of information to effectively mimic task behaviours. Currently, Decision Transformer (DT) and subsequently, the Hierarchical Decision Transformer (HDT), presented Transformer-based approaches to learn task policies. Recently, the Mamba architecture has shown to outperform Transformers across various task domains. In this work, we introduce two novel methods, Decision Mamba (DM) and Hierarchical Decision Mamba (HDM), aimed at enhancing the performance of the Transformer models. Through extensive experimentation across diverse environments such as OpenAI Gym and D4RL, leveraging varying demonstration data sets, we demonstrate the superiority of Mamba models over their Transformer counterparts in a majority of tasks. Results show that DM outperforms other methods in most settings. 
\end{enumerate}

\section{Related Work}
\label{sec:relatedwork}

Offline RL is a machine learning paradigm that allows agents to learn tasks from pre-existing demonstration data sets collected by a set of policies and has been applied to multiple domains such as playing games, driving autonomously and robotics \cite{senhoras}.
In this section, we provide a summary of some works related to ours.
BC treats offline RL as a direct supervised learning problem, where the problem aims to copy the state-action maps represented in the data set, where the training signal is given by how similar the actions are to the demonstrator's \cite{mario}.

Sequence modelling with deep networks has evolved from LSTMs to Transformer architectures \cite{attention}. Due to the self-attention mechanism, Transformers have revolutionised many natural language processing tasks. 
Recently, they have been applied to RL \cite{dt,tt}, by re-framing it as a sequence modelling problem. In the DT, the agent is conditioned on past trajectories and the accumulated reward to be collected in the future, the returns-to-go (RTG). Although the DT has shown success in many tasks, it has been proven to be reliant on the sequence of RTG to perform \cite{hdt}. This reliance hinders its performance in stochastic environments, where the reward sequence isn't deterministic and requires a user to specify the appropriate desired reward for the specific task. The latter is non-trivial and affects the performance.

Since then, a few methods have been proposed to tackle issues with the DT. In \cite{onlinedt}, online learning is used to train the Transformer. Alternatively, \cite{transpretrain} pre-trains the Transformer on large corpus of text which in turn increases performance on seemingly unrelated tasks. To tackle the DT's dependency on the RTG, a few methods have proposed ways to replace this sequence. In \cite{td}, a value function is trained beforehand using the demonstration data set. Then, the sequence of RTG is replaced with the state value predictions of the estimated reward function. Although the method tackles the stochasticity problem, it introduces the reliance on a quality value function which is limited by the demonstration data set.
Other works target the stochasticity problem of DT. The method in \cite{splt}, aims to estimate environmental stochasticity using a Transformer model to aid policy learning of the main Transformer.
Alternatively, hierarchical approaches learn a high-level planner and a low-level controller \cite{iris,swirl}.
Here, the high-level model divides the problem into smaller tasks and conditions the low-level model on sub-goals that lead it to achieving the main goal.
Conditioning reinforcement learning and imitation learning approaches on goal observations improves sample efficiency.
HDT \cite{hdt} proposes to identify sub-goal states in the demonstration trajectories, and train a high-level method to predict such sub-goals based on the current state trajectory. Then, the low level DT is conditioned by the sequence of predicted sub-goals instead of RTG. However, these methods required an additional model to replace the sequence of RTG. 

Although Transformers have achieved impressive capabilities due to their self-attention mechanism, their scalability is constrained by quadratic scaling relative to the size of the context window. In contrast, structured state space sequence models (SSMs) \cite{ssms} have gained attention for their linear scalability with the sequence length. Notably, the Mamba architecture \cite{mamba} merges the context-dependent reasoning of Transformers with the linear scalability of SSMs through its selection mechanism. Mamba has demonstrated superiority over Transformers in numerous sequence processing tasks \cite{hiss}. Hence, we advocate for replacing Transformers with the Mamba architecture in both HDT and DT models. We demonstrate that the evolutionary parameter of SSMs can effectively substitute the sequence of rewards in DT, guiding the agent through the task.

\section{Preliminaries}
\label{sec:preliminaries}

\subsection{Reinforcement Learning}

Reinforcement learning can be defined as a Markov Decision Process (MDP) described by the tuple $(S, A, P, R)$. Where $S$ is the set of states, $A$ is the set of actions, $P(s'\mid s, a)$ is the state transition function and $R(s, a)$ is the reward function. We use $s_t$, $a_t$, and $r_t = R(s_t, a_t)$ to denote the state, action, and reward at time step $t$, respectively. 
At every time step $t$, the agent observes the environment's state $s_t$ and selects an action based on its current policy $a_t = \pi(s_t)$.
The agent then performs the action, obtains a reward $r_t = R(s_t, a_t)$ for the interaction, and transitions to the next environment state $s_{t+1} \sim P(s_{t+1}\mid s_t, a_t)$.
A trajectory is a sequence of length $N$ of states, actions, and rewards: $\tau = (s_1, a_1, r_1, ... , s_N , a_N , r_N )$. The cumulative rewards of a trajectory $R_\tau$ with length $N$ are: $\sum_{t=1}^{N} r_t$. 
The goal of RL is to estimate the policy function $\pi$ that produces trajectories $\tau$ which maximize the expected return $\mathop{\mathbb{E}}_\pi [R_\tau]$.

\subsection{Offline RL}

In offline RL, the agent has access to a demonstration data set $D = \{\tau_1, ..., \tau_N\}$ collected by a set of policies. Instead of learning through trial and error interactions with the environment, the agent learns solely from the trajectories present in the demonstration data set. 

The simplest form of offline RL is through Behaviour Cloning (BC). In BC the agent is encouraged to directly imitate the demonstrator, by selecting the actions the demonstrator took for every single state in the data set. A common approach is to maximize the likelihood of actions in the demonstration, $\max\mathop{\mathbb{E}}_{(s,a)\sim D} log \pi(a\mid s)$. 
Although the agent remains safe while learning, the downside of this setting is that the agent is dependent on the quality and size of the data set, and blindly copying the actions can lead to compounding errors. This latter problem can be alleviated by using a sequence model. By providing the agent with a sequence of transitions instead of a single observation, sequence models have surpassed linear models in BC, by reducing the compounding error caused by learning from limited data.

\subsection{Decision Transformer}

DT formulates RL as a sequence objective problem and applies the Transformer architecture because of their success in a wide range of applications. Specifically, the DT employs the GPT 2 \cite{gpt} architecture. Instead of processing a single state observation, the policy selects the next action, based on a sequence of states, actions and returns-to-go (RTG): $(rtg_1, s_1, a_1, ... , rtg_N, s_N, a_N)$. During training, the RTG are the remaining cumulative return obtained in the trajectory after time step $t$: $rtg_t = \sum_{t'=t}^{T} r_{t'}$. However, during deployment, the full trajectory is not known before execution to determine the initial value of the sequence. Instead, this initial value of desired returns must be specified by the user, which is task-specific and affects the performance \cite{hdt}. The original DT sets the value to the maximum accumulated rewards found in the data set.

The DT also requires a causal mask, which is a binary vector with as many elements as the length of the sequence. Each element in the mask determines if the corresponding tokens should be hidden from the Transformer to produce the output. Additionally, the Transformers require information about the relative position of the tokens in the sequence. For the DT, the positional encoding is done by creating a sequence of time steps, passing this sequence through an embedding layer and summing the result to each of the three token sequences.
The DT is trained using the L2 loss:

\begin{equation}
\resizebox{0.9\linewidth}{!}{$\mathop{\mathbb{L}} (s_{t:t+K}, a_{t:t+K}, rtg_{t:t+K}) = \|a_{t+K} - \pi_{\phi} (s_{t:t+K}, a_{t:t+K-1}, rtg_{t:t+K})\|$}
\end{equation}

\subsection{Hierarchical Decision Transformer}
\label{sec:prel_hdt}

The HDT tackles the reliance of the DT on the sequence of RTG.
It augments the original MDP with sets of absorbing goal and sub-goal states $G \subset S$ and $Sg \subset S$. Where each goal state $g \in G$ is a state of the world in which the task is considered to be solved and $sg \in Sg$ is a valuable state of the world that contributes to the success of the trajectory.
It then replaces the sequence of RTG with a sequence of sub-goal states. Before training, the data set is processed, where each transition in the data set is augmented with the highest valued state from the remaining trajectory. The value of each following state in the trajectory is determined by: $W(s_j) = \sum_{k=i+1}^{j} \frac{r_k}{j-i}$.
The result is a data set of $M$ trajectories $D = \{\tau_1,..., \tau_M\}$, where each trajectory $\tau_j = \{(s_i, a_i, sg_i), i \in N\}$, where N is the length of the trajectory.

The HDT splits the decision-making process of the agent into two models: a high-level mechanism which defines sub-goal states for a low-level controller to try and reach.
The high-level mechanism receives the sequences of past states and sub-goals and then aims to produce the next sub-goal for the low-level controller to reach, guiding it through the task.
The low-level controller receives the sequences of past states, actions and sub-goals to predict the next correct action.
Similarly to the DT, both models also receive the sequence of time steps for positional encoding, as well as the causal mask, and are trained with the L2 loss.

\subsection{Structured State Space Sequence Models}

SSMs and Mamba model continuous systems that map a 1-D function $x(t) \xrightarrow{} y(t) \in \mathbb{R}$ through a hidden state $h(t) \in \mathbb{R}^N$. This process can be represented as a linear Ordinary Differential Equation (ODE):

\begin{equation}
\begin{aligned}
  h'(t) = \textbf{A}h(t) + \textbf{B}x(t),\ \ \ \ y(t) = \textbf{C}h(t)
\end{aligned}
\end{equation}

where, $\textbf{A} \in \mathbb{R}^{N \times N}$ serves as the evolution parameter, while $\textbf{B} \in \mathbb{R}^{N \times 1}$ and $\textbf{C} \in \mathbb{R}^{N \times 1}$ act as the projection parameters.

S4 models adapt continuous systems for deep learning applications through discretisation of the true continuous function. They introduce a timescale parameter, $\Delta$ which determines the precision, and then convert the continuous parameters $\textbf{A}$ and $\textbf{B}$ into discrete parameters $\bar{\textbf{A}}$ and $\bar{\textbf{B}}$. There are different ways to perform discretization, but the original methods choose to use the zero-order hold (ZOH) method:

\begin{equation}
\begin{aligned}
  \bar{\textbf{A}} = \exp(\Delta\textbf{A}),  \ \ \ \
  \bar{\textbf{B}} = (\Delta\textbf{A})^{-1}(\exp(\Delta\textbf{A}) - \textbf{I}).\Delta\textbf{B}
\end{aligned}
\end{equation}

After the discretisation, the models can be rewritten as:

\begin{equation}
\begin{aligned}
  h'(t) = \bar{\textbf{A}}h(t) + \bar{\textbf{B}}x(t),  \ \ \ \ y(t) = \textbf{C}h(t)
\end{aligned}
\end{equation}

Lastly, the models compute the output through a global convolution:

\begin{equation}
\begin{aligned}
  \bar{\textbf{K}} = (\textbf{C}\bar{\textbf{B}}, \textbf{C}\bar{\textbf{A}}\bar{\textbf{B}}, ..., \textbf{C}\bar{\textbf{A}}^{K-1}\bar{\textbf{B}}),\ \ \ \ y(t) = x * \bar{\textbf{K}}
\end{aligned}
\end{equation}

where $\bar{\textbf{K}} \in \mathbb{R}^K$ represents a structured convolutional kernel, and $K$ denotes the length of the input sequence $x$.

\section{Methodology}

We propose to replace the Transformer architecture of both the and HDT methods with the Mamba architecture, resulting in the DM and the HDM. This substitution, simplifies the overall architecture of the methods by removing the need for a causal mask and for positional encoding. Furthermore, for DM this causes the model to no longer rely on the sequence of RTG to perform the task, unlike the DT. We further explain both the DM and the HDM in the following sub-sections. The architectures of DM and HDM are represented in Fig. \ref{fig:hdm}.

\subsection{Decision Mamba}

We assume to have access to a demonstration data set $D$ composed of trajectories with states, actions and rewards. Similarly to the DT, we condition the model on the sequences of states actions and RTG, each sequence with a context length of $K$. However, we perform experiments that show that the DM is also able to perform without the sequence of RTG, while improving the performance. 
During batch data sampling, a demonstration trajectory is randomly selected, followed by the choice of a trajectory index $t \in [0, T - 1]$, where $T$ is the length of the trajectory.
The $K-1$ states, actions and RTG preceding $t$ compose the three sequences. We also sample a right-shifted sequence of actions. This sequence corresponds to the ground-truth actions that the model should predict. The sequences are then padded with zero-filled vectors to the right, if their length is smaller than the context length. Because of this padding, we create a binary padding mask. This is a sequence of length $K$ with zeros in indexes where the previous sequences were padded. This padding mask is solely used to prevent the loss function from considering padded values, and is not used by the DM model or during inference.

Each of the three input sequences pass through individual linear layers, converting them to the embedding dimension used by the model.
The resulting embeddings are joined to form the sequential input of the model in an identical way to what is done in DT. The difference is that the DM does not require the causal mask of the Transformer, and Mamba provides sequential information. Therefore, we can skip a significant number of steps. We don't need to sample an additional sequence time steps from the data set, that would be used to generate a positional encoding, and this encoding would then be summed to the three embeddings.

DM employs a Mamba architecture with repeating Mamba layers each receiving the embeddings of the previous layer and outputting new embedding vectors.
We perform tests varying the number of layers, the embedding size, and the context length. There wasn't an overall configuration that resulted in superior performance across all the tasks. In the experiments section, we use 6 Mamba layers, an embedding size of 128, and a context length of 20.
The embeddings returned by the final Mamba layer pass through a feed-forward layer to project them to the action space of the task. Subsequently, a hyperbolic tangent activation function normalises the values between -1 and 1, which are then multiplied by the respective action range of the task.

The DM's objective is to predict the corresponding ground-truth actions in the data set after being conditioned on the sequence of past states, actions and optionally RTG. Because of this, the model's objective is the L2 loss between the predicted action sequence and the ground-truth action sequence.



\subsection{Hierarchical Decision Mamba}

For the HDM, we follow the same data processing pipeline of HDT and replace the sequences of RTG with sequences of sub-goals as detailed in Sec. \ref{sec:prel_hdt}. The HDM is also composed of two models: the high-level mechanism and low-level controller. Similarly to the DM, during batch sampling we sample padded sequences of states, actions and sub-goals of length $K$ from the data set and create the padding mask. The high-level mechanism receives the sequences of states and sub-goals and predicts new sub-goals, while the low-level controller receives the sequences of states, actions and sub-goals and predicts new actions. Therefore, we also sample right-shifted sequence of actions and right-shifted sequence of sub-goals, as the ground-truth for the low-level, and high-level model respectively.

The sequences of states and sub-goals each pass through a linear layer, converting them to the embedding dimension used by the model. They are then joined and passed to the Mamba layers of the high-level mechanism. The output vectors of the final Mamba layer are passed to a linear layer which projects them to the state space of the task. Then, a hyperbolic tangent activation function normalises the values between -1 and 1. Following the HDT, the states of the data set are normalised between -1 and 1. The high-level mechanism is trained using the L2 loss between the predicted sub-goal sequence and the ground-truth sub-goal sequence.

The training of the low-level controller is similar to the training of the DM. Instead of receiving a sequence of RTG, it receives a sequence of sub-goals. The low-level controller is trained using the L2 loss between the predicted sequence of actions and the ground-truth action sequence.

\section{Experiments}
\label{sec:experiments}

\begin{figure*}[!tb]
\centering
\includegraphics[width=1.0\linewidth]{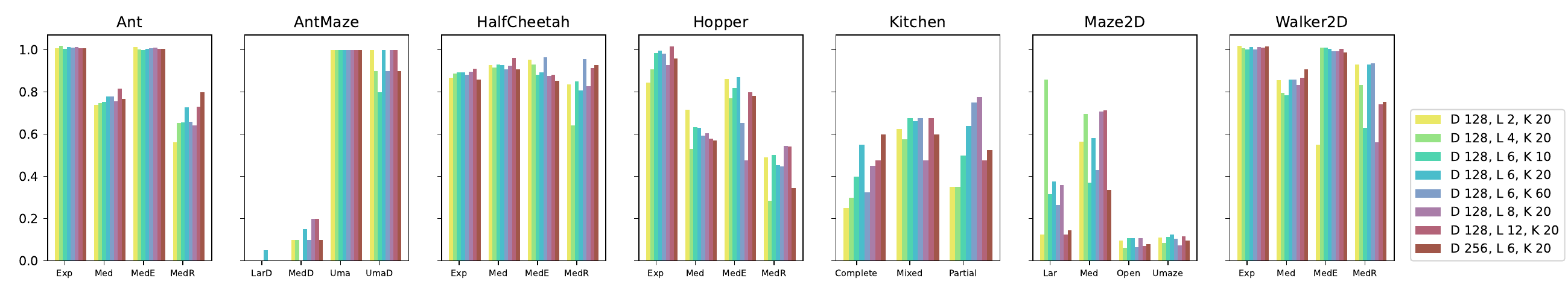}
\caption{Comparison of the performance of the HDM varying architecture configuration across the 7 D4RL tasks, for different demonstration data sets. The scale of the bar graphs is the maximum reward present in the respective data set. L is the number of layers, D is the embedding size, and K is the context length.}
\label{fig:hdm_graphs}
\end{figure*}

\begin{figure*}[!tb]
\centering
\includegraphics[width=1.0\linewidth]{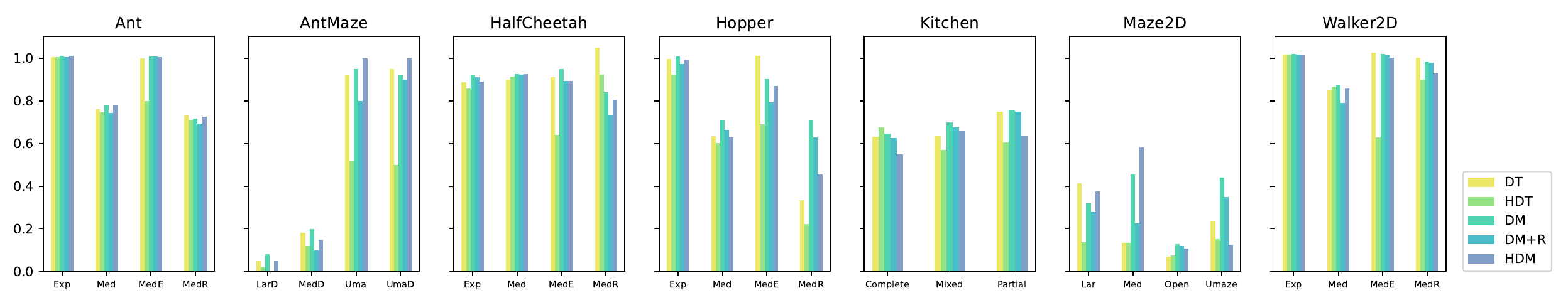}
\caption{Comparison of the 5 methods across the 7 D4RL tasks, for different the demonstration data sets. The scale of the bar graphs is the maximum reward present in the respective data set. All models have 6 layers, an embedding size of 128 and use context length of 20. The values of DM and DT are obtained by using the maximum reward of the data set as the desired reward.}
\label{fig:all_graphs}
\end{figure*}

\begin{figure*}[!tb]
\centering
\includegraphics[width=1.0\linewidth]{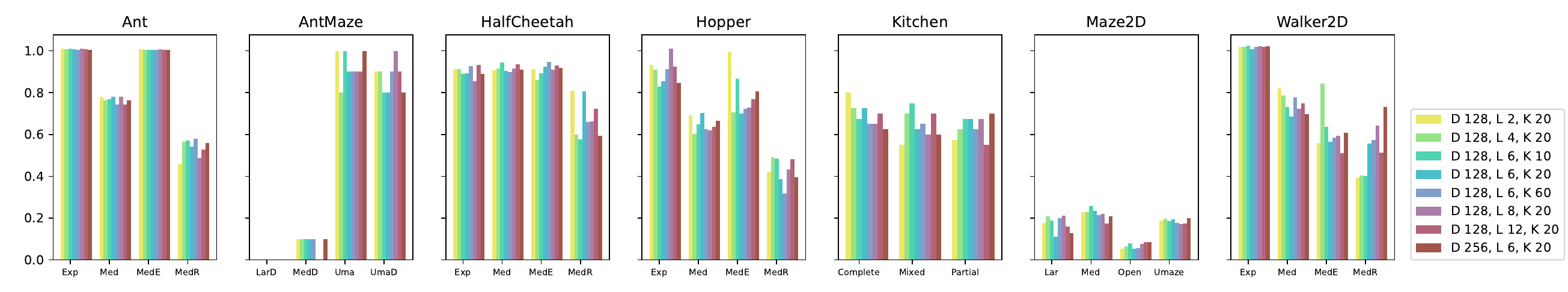}
\caption{Comparison of the performance of the DM varying architecture configuration across the 7 D4RL tasks, for different demonstration data sets. The scale of the bar graphs is the maximum reward present in the respective data set. L is the number of layers, D is the embedding size, and K is the context length.}
\label{fig:dm_nor_graphs}
\end{figure*}

\begin{figure*}[!tb]
\centering
\includegraphics[width=1.0\linewidth]{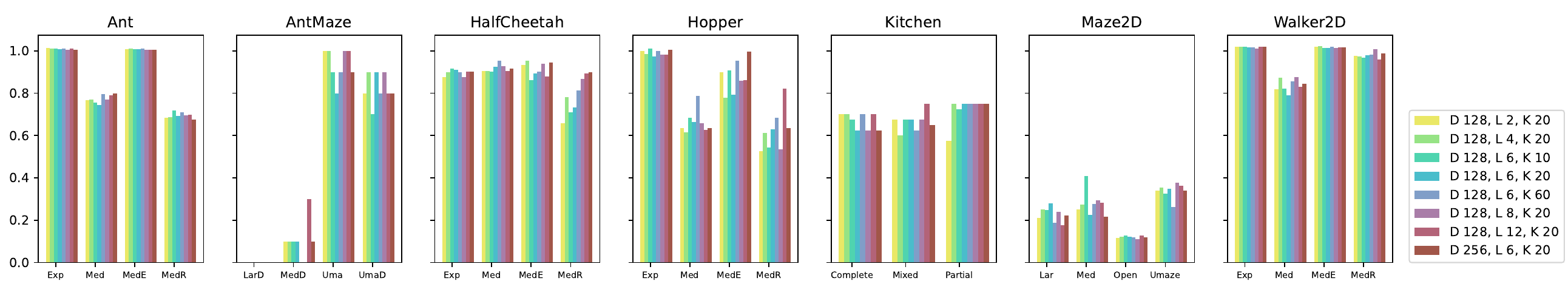}
\caption{Comparison of the performance of the DM with the sequence of RTG, varying architecture configuration across the 7 D4RL tasks, for different demonstration data sets. The scale of the bar graphs is the maximum reward present in the respective data set. L is the number of layers, D is the embedding size, and K is the context length. The values are obtained by using the maximum reward of the data set as the desired reward.}
\label{fig:dm_graphs}
\end{figure*}

\begin{table*}[!tb]
\caption{Maximum accumulated returns of the DT, HDT, DM, DM with RTG, and HDM methods using 6 layers, an embedding size of 128 and context length of 20. We test the models on seven tasks from the D4RL \cite{d4rl} benchmark and vary the demonstration data sets. Highest values are highlighted in bold.}
\label{tab:res}
\resizebox{\linewidth}{!}{
\begin{tabular}{cc|ccc|c|ccc|c|c}
\multirow{2}{*}{\textbf{Task}} & \multirow{2}{*}{\textbf{Data Set}} & \multicolumn{3}{c|}{\textbf{DT}}                                             & \multirow{2}{*}{\textbf{HDT}} & \multicolumn{3}{c|}{\textbf{DM w/ R}}                                       & \multirow{2}{*}{\textbf{DM wo/ R}} & \multirow{2}{*}{\textbf{HDM}} \\
                               &                                    & Half               & Max                        & 10k                        &                               & Half               & Max                        & 10k                       &                                    &                               \\ \hline
\multirow{4}{*}{Ant}           & expert                             & 2722.84 $\pm$ 9.27   & 2731.52 $\pm$ 3.91           & 2733.84 $\pm$ 9.3            & 2731.83 $\pm$ 8.38              & 1853.28 $\pm$ 7.54   & 2735.34 $\pm$ 11.45          & 2742.30 $\pm$1 14.37        & 2746.57 $\pm$ 10.69                  & \textbf{2749.2 $\pm$ 11.62}     \\
                               & medium                             & 787.54 $\pm$ 22.87   & 801.41 $\pm$ 14.61           & 800.91 $\pm$ 17.32           & 787.23 $\pm$ 33.33              & 760.23 $\pm$ 21.73   & 782.97 $\pm$ 13.83           & 803.63 $\pm$ 14.76          & \textbf{821.03 $\pm$ 18.21}          & 820.39 $\pm$ 23.41     \\
                               & medium-expert                      & 1525.61 $\pm$ 724.43 & 2718.68 $\pm$ 11.66          & 2713.55 $\pm$ 15.06          & 2171.27 $\pm$ 792.37            & 1769.70 $\pm$ 18.46  & \textbf{2742.88 $\pm$ 13.64} & 2734.04 $\pm$ 15.92         & 2738.41 $\pm$ 8.42                   & 2730.01 $\pm$ 12.45             \\
                               & medium-replay                      & 650.23 $\pm$ 63.67   & 732.89 $\pm$ 13.76           & \textbf{815.53 $\pm$ 42.72}  & 712.61 $\pm$ 39.5               & 639.22 $\pm$ 20.49   & 694.40 $\pm$ 13.69           & 716.67 $\pm$ 37.54          & 716.85 $\pm$ 33.65                   & 728.2 $\pm$ 48.64               \\ \hline
\multirow{4}{*}{Antmaze}       & large-diverse                      & 0.02 $\pm$ 0.04      & 0.05 $\pm$ 0.05              & 0.05 $\pm$ 0.05              & 0.02 $\pm$ 0.04                 & 0.0 $\pm$ 0.01       & 0.0 $\pm$ 0.01               & 0.0 $\pm$ 0.01              & \textbf{0.08 $\pm$ 0.13}             & 0.05 $\pm$ 0.05                 \\
                               & medium-diverse                     & 0.2 $\pm$ 0.17       & 0.18 $\pm$ 0.19              & 0.2 $\pm$ 0.17               & 0.12 $\pm$ 0.04                 & 0.0 $\pm$ 0.05       & 0.1 $\pm$ 0.05               & \textbf{0.5 $\pm$ 0.05}     & 0.2 $\pm$ 0.17                       & 0.15 $\pm$ 0.05                 \\
                               & umaze                              & 0.92 $\pm$ 0.04      & 0.92 $\pm$ 0.04              & 0.92 $\pm$ 0.04              & 0.52 $\pm$ 0.25                 & 0.8 $\pm$ 0.04       & 0.8 $\pm$ 0.04               & 0.9 $\pm$ 0.04              & 0.95 $\pm$ 0.05                      & \textbf{1.0 $\pm$ 0.0}          \\
                               & umaze-diverse                      & 0.92 $\pm$ 0.04      & 0.95 $\pm$ 0.05              & 0.9 $\pm$ 0.07               & 0.50 $\pm$ 0.25                 & 0.9 $\pm$ 0.04       & 0.9 $\pm$ 0.07               & 0.9 $\pm$ 0.05              & 0.92 $\pm$ 0.04                      & \textbf{1.0 $\pm$ 0.0}          \\ \hline
\multirow{4}{*}{HalfCheetah}   & expert                             & 1516.19 $\pm$ 25.85  & 1530.86 $\pm$ 21.28          & 1497.99 $\pm$ 48.14          & 1479.88 $\pm$ 22.09             & 1025.66 $\pm$ 23.74  & 1570.21 $\pm$ 27.80          & 1573.29 $\pm$ 46.12         & \textbf{1585.85 $\pm$ 19.24}         & 1538.32 $\pm$ 50.83             \\
                               & medium                             & 655.96 $\pm$ 8.55    & 658.54 $\pm$ 2.02            & 667.08 $\pm$ 16.17           & 669.73 $\pm$ 15.74              & 597.75 $\pm$ 7.69    & 676.78 $\pm$ 3.14            & 664.15 $\pm$ 15.39          & 677.84 $\pm$ 13.6                    & \textbf{679.03 $\pm$ 14.74}     \\
                               & medium-expert                      & 1432.58 $\pm$ 158.2  & 1572.42 $\pm$ 68.04          & 1619.32 $\pm$ 43.02          & 1103.51 $\pm$ 141.89            & 1097.76 $\pm$ 68.47  & 1541.82 $\pm$ 58.53          & 1566.91 $\pm$ 83.36         & \textbf{1636.29 $\pm$ 27.28}         & 1541.19 $\pm$ 73.06             \\
                               & medium-replay                      & 806.58 $\pm$ 175.57  & 1074.72 $\pm$ 11.14          & \textbf{1100.29 $\pm$ 15.2}  & 946.68 $\pm$ 91.13              & 607.06 $\pm$ 110.68  & 749.42 $\pm$ 13.93           & 916.15 $\pm$ 67.71          & 862.53 $\pm$ 35.02                   & 826.06 $\pm$ 96.8               \\ \hline
\multirow{4}{*}{Hopper}        & expert                             & 2129.4 $\pm$ 116.64  & 2234.25 $\pm$ 54.94          & 2239.86 $\pm$ 58.55          & 2072.71 $\pm$ 69.58             & 1211.95 $\pm$ 123.71 & 2183.71 $\pm$ 58.71          & 2218.42 $\pm$ 64.48         & \textbf{2260.68 $\pm$ 22.36}         & 2230.41 $\pm$ 65.08             \\
                               & medium                             & 1144.2 $\pm$ 100.69  & 1241.6 $\pm$ 188.58          & 1252.95 $\pm$ 99.69          & 1175.3 $\pm$ 63.39              & 1101.02 $\pm$ 97.47  & 1299.97 $\pm$ 143.74         & 1289.51 $\pm$ 89.36         & \textbf{1383.51 $\pm$ 127.34}        & 1231.19 $\pm$ 108.38            \\
                               & medium-expert                      & 2056.43 $\pm$ 166.75 & \textbf{2270.67 $\pm$ 62.89} & 2259.2 $\pm$ 117.52          & 1549.4 $\pm$ 112.42             & 1128.34 $\pm$ 128.57 & 1776.79 $\pm$ 86.91          & 1914.85 $\pm$ 105.72        & 2026.48 $\pm$ 180.21                 & 1948.04 $\pm$ 342.99            \\
                               & medium-replay                      & 636.21 $\pm$ 91.45   & 564.7 $\pm$ 99.69            & 648.13 $\pm$ 172.74          & 378.57 $\pm$ 176.15             & 822.5 $\pm$ 86.30    & 1065.73 $\pm$ 90.73          & 1175.59 $\pm$ 95.76         & \textbf{1199.42 $\pm$ 192.37}        & 770.51 $\pm$ 95.49              \\ \hline
\multirow{3}{*}{Kitchen}       & complete                           & 2.52 $\pm$ 0.18      & 2.53 $\pm$ 0.13              & 2.42 $\pm$ 0.38              & 2.7 $\pm$ 0.23                  & \textbf{3.1 +0.16} & 2.5 $\pm$ 0.09               & 3.0 $\pm$ 0.02              & 2.58 $\pm$ 0.36                      & 2.2 $\pm$ 0.14                  \\
                               & mixed                              & 2.28 $\pm$ 0.27      & 2.55 $\pm$ 0.21              & 2.15 $\pm$ 0.34              & 2.28 $\pm$ 0.13                 & 2.5 $\pm$ 0.26       & 2.7 $\pm$ 0.18               & 2.5 $\pm$ 0.37              & \textbf{2.8 $\pm$ 0.2}               & 2.65 $\pm$ 0.09                 \\
                               & partial                            & 2.65 $\pm$ 0.27      & 3.0 $\pm$ 0.07               & 2.08 $\pm$ 0.51              & 2.42 $\pm$ 0.44                 & 2.1 $\pm$ 0.25       & 3.0 $\pm$ 0.03               & 3.0 $\pm$ 0.02              & \textbf{3.02 $\pm$ 0.04}             & 2.55 $\pm$ 0.43                 \\ \hline
\multirow{4}{*}{Maze2D}        & large                              & 106.72 $\pm$ 17.23   & 103.7 $\pm$ 15.03            & 103.45 $\pm$ 19.85           & 34.1 $\pm$ 8.97                 & 32.7 $\pm$ 16.05     & 69.9 $\pm$ 14.08             & \textbf{110.8 $\pm$ 17.74}  & 79.65 $\pm$ 20.43                    & 93.7 $\pm$ 71.77                \\
                               & medium                             & 40.07 $\pm$ 9.19     & 33.72 $\pm$ 13.3             & \textbf{154.0 $\pm$ 63.45}   & 33.72 $\pm$ 7.87                & 37.8 $\pm$ 5.34      & 56.3 $\pm$ 5.35              & 111.9 $\pm$ 45.34           & 114.1 $\pm$ 75.94                    & 145.23 $\pm$ 20.81              \\
                               & open                               & 16.62 $\pm$ 0.48     & 17.17 $\pm$ 1.49             & 15.85 $\pm$ 3.61             & 19.08 $\pm$ 2.17                & 25.7 $\pm$ 1.64      & 30.2 $\pm$ 8.63              & \textbf{32.4 $\pm$ 4.56}    & 32.33 $\pm$ 2.05                     & 26.67 $\pm$ 0.76                \\
                               & umaze                              & 58.68 $\pm$ 20.14    & 59.3 $\pm$ 20.61             & 61.2 $\pm$ 19.74             & 37.72 $\pm$ 11.7                & 58.8 $\pm$ 17.03     & 86.9 $\pm$ 10.47             & \textbf{186.1 $\pm$ 16.43}  & 110.25 $\pm$ 51.04                   & 31.13 $\pm$ 2.58                \\ \hline
\multirow{4}{*}{Walker2D}      & expert                             & 1863.25 $\pm$ 11.68  & 1867.61 $\pm$ 9.43           & 1877.92 $\pm$ 5.12           & 1869.07 $\pm$ 8.54              & 1088.01 $\pm$ 14.72  & 1866.81 $\pm$ 4.72           & \textbf{1881.46 $\pm$ 9.46} & 1874.51 $\pm$ 10.86                  & 1861.86 $\pm$ 7.2               \\
                               & medium                             & 1070.96 $\pm$ 61.03  & 1081.54 $\pm$ 42.68          & \textbf{1175.83 $\pm$ 58.49} & 1106.63 $\pm$ 24.09             & 932.2 $\pm$ 40.81    & 1008.24 $\pm$ 54.91          & 1088.64 $\pm$ 43.09         & 1111.94 $\pm$ 32.08                  & 1093.42 $\pm$ 36.83             \\
                               & medium-expert                      & 1310.26 $\pm$ 352.94 & \textbf{1883.23 $\pm$ 19.77} & 1880.87 $\pm$ 24.98          & 1153.5 $\pm$ 54.57              & 1055.61 $\pm$ 35.89  & 1861.45 $\pm$ 23.23          & 1866.47 $\pm$ 18.67         & 1871.34 $\pm$ 4.88                   & 1843.48 $\pm$ 11.49             \\
                               & medium-replay                      & 1129.27 $\pm$ 129.82 & 1335.64 $\pm$ 10.87          & \textbf{1377.9 $\pm$ 21.19}  & 1197.66 $\pm$ 51.96             & 716.23 $\pm$ 78.38   & 1302.54 $\pm$ 12.94          & 1330.53 $\pm$ 34.71         & 1310.88 $\pm$ 32.54                  & 1237.97 $\pm$ 63.21            
\end{tabular}
}
\end{table*}

In this section, we assess the performance of the proposed DM and HDM models, comparing them with Transformer-based methods, DT and HDT. Our evaluation spans over seven distinct tasks sourced from the D4RL benchmark, selected precisely for its provision of a diverse set of tasks, each accompanied by multiple demonstration data sets of varying quality.
As the efficacy of these methods relies heavily on their capacity to approximate policies represented in the demonstration data set, the resulting performance of the methods is intrinsically linked to the quality of these data sets. Consequently, we conduct our training procedures using the diverse data sets provided by the benchmark for each respective task. 

We train each model for 1 million epochs, using batch sizes of 16, and a learning rate of $1e^{-4}$.
To mitigate the influence of outliers and the inherent seed-dependency of episodes, we adopt a validation strategy wherein, every one thousand epochs, we validate the model on 100 episodes, and compute the average accumulated rewards. 
Due to the seed dependency, we repeat each experiment across 4 different random seeds.
The presented results are the average values across the 4 seeds of the highest accumulated rewards seen throughout the 1 million epochs.

We also varied the number of Mamba and Transformer layers, the embedding size inside the model, and the length of the token sequence. Results for DM, DM with RTG, and HDM are shown in Fig. \ref{fig:hdm_graphs}, Fig. \ref{fig:dm_nor_graphs} and Fig. \ref{fig:dm_graphs}, respectively. The values of the DM with RTG were obtained using a desired reward equal to the maximum reward present in the respective demonstration data set. However, results were unclear, as there wasn't a best performing configuration for any of the models. Different configurations resulted in better or worse performance for the models on the different tasks and data sets. To fairly compare the models using the same architecture, we chose the architecture that better represented the average results across the set of architectures. This architecture is composed of 6 layers (Transformer or Mamba), an embedding size of 128 and a sequence length of 20.
For the execution of DT and DM using the sequence of RTG, we initially identify the maximum accumulated returns attained by a trajectory in the demonstration data set. Subsequently, during validation, we set the desired returns to this maximum value, half of it, and a larger value—specifically, 10k. The results are presented in Table \ref{tab:res}.

One of the driving motivations behind the development of the HDT was to alleviate the necessity of manually specifying the desired RTG, a notable challenge encountered in the evaluation and deployment of DTs.
To determine whether the DM inherits this drawback from DT, we evaluate whether it also relies on the RTG sequence to guide the model, or if the sequence can be simply removed from the model's input. Table \ref{tab:res} presents the accumulated returns achieved by the DM model without the desired returns sequence, compared with a variant of the DM model with this additional sequence. 

The results depicted in the table highlight that the DM does not require the sequence of RTG for effective performance.
Additionally, results also show that the DM without the sequence of RTG does seem to reach a slightly better performance than with the sequence. 
This is also true across different architectures as shown in Fig. \ref{fig:dm_nor_graphs}. Notably, in the DT, eliminating this sequence impedes the DT's ability to learn the task entirely as shown in \cite{hdt}. This indicates that the evolutionary parameter of the Mamba architecture successfully replaces the need for RTG.
Since DM without the sequence of rewards achieves higher performance without requiring additional user interaction, we can conclude that DM should be used without the reward sequence. 

Lastly, we compare the new proposed Mamba methods with their Transformer predecessors. According to the results in Table \ref{tab:res}, the DM without rewards outperforms the DT in 15 out of the 27 settings. Additionally, the HDM outperforms the HDT in 22 out of the 27 settings. The superiority of the Mamba models exists even while comparing to the DT with the ideal desired reward for each task. When comparing the DM to the DT using a fixed desired reward of 10k, the DM outperforms the DT in 17 out of the 27 settings. These results show that the proposed Mamba methods improve upon the Transformer predecessors in the D4RL benchmark.
Overall, the DM without rewards is the best performing model of the set. 
Although, the HDM and the DT are still very competitive, it is worth noting that HDM requires two models and pre-processing the data set, while DT requires user interaction and task knowledge. Moreover, unlike the other models, DM can be applied to tasks without a reward function.

\subsection{Time Comparison}

Mamba models have outpaced Transformers in terms of speed in other applications. Also, HDT and HDM require the training of two models, and the extra computational cost may not be worth the performance benefits.
Because of this, we compare the time required to perform a training iteration and the inference step using the different methods across the 7 task environments available in the D4RL benchmark. We use a batch size of 16, an embedding size of 128, a sequence length of 20, and 6 layers per model. Specifically, for the HDT and the HDM, we employ 6 layers for both the high-level and low-level models, and we measure the time to train both models. For training, we measure the time it takes for a gradient calculation and update. For inference, we measure the time it takes to build the sequences, obtain an action from the model and perform the transition. To ensure statistical robustness, we repeat both these steps 1000 times for each model, presenting the average and standard deviation time to perform a training iteration, and an inference step in Table \ref{tab:time}. Results show that during training there's not much difference between the Mamba methods and their Transformer predecessors. As expected the HDT and the HDM take close to double the time to train due to having double the models than the DT and the DM, respectively. Adding rewards to the DM does not increase the training time significantly. At inference time however, the Mamba methods are faster than the Transformer methods. In addition to the increase in performance, this computational boost further shows the benefits of our methods compared to the baselines.

\begin{table*}[tb]
\caption{Average and STD time for a single training iteration, and to perform inference of an episode, across the different D4RL tasks, using a batch size of 16, of each of the 5 methods, configured with 6 layers, an embedding size of 128 and sequence length 20. Lowest values are highlighted in bold.}
\label{tab:time}
\resizebox{\linewidth}{!}{
\begin{tabular}{c|ccccc|lllll}
\multicolumn{1}{l|}{} & \multicolumn{5}{c|}{\textbf{Train Time (s)}}                                                                                                                                   & \multicolumn{5}{c}{\textbf{Inference Time (s)}}                                                                                                                                                  \\ \cline{2-11} 
\textbf{}             & \multicolumn{1}{c|}{\textbf{DT}}     & \multicolumn{1}{c|}{\textbf{HDT}}    & \multicolumn{1}{c|}{\textbf{DM wo/ R}} & \multicolumn{1}{c|}{\textbf{DM w/ R}} & \textbf{HDM}    & \multicolumn{1}{c|}{\textbf{DT}}     & \multicolumn{1}{c|}{\textbf{HDT}}    & \multicolumn{1}{c|}{\textbf{DM wo/ R}} & \multicolumn{1}{c|}{\textbf{DM w/ R}} & \multicolumn{1}{c|}{\textbf{HDM}} \\ \hline
Ant                   & \multicolumn{1}{c|}{\textbf{0.015 $\pm$ 0.011}} & \multicolumn{1}{c|}{0.020 $\pm$ 0.011} & \multicolumn{1}{c|}{0.018 $\pm$ 0.102}   & \multicolumn{1}{c|}{0.018 $\pm$ 0.097}  & 0.026 $\pm$ 0.101 & \multicolumn{1}{l|}{0.005 $\pm$ 0.007} & \multicolumn{1}{l|}{0.007 $\pm$ 0.001} & \multicolumn{1}{l|}{\textbf{0.003 $\pm$ 0.004}}   & \multicolumn{1}{l|}{\textbf{0.003 $\pm$ 0.004}}  & 0.005 $\pm$ 0.008                   \\
Antmaze               & \multicolumn{1}{c|}{0.008 $\pm$ 0.000} & \multicolumn{1}{c|}{0.014 $\pm$ 0.000} & \multicolumn{1}{c|}{\textbf{0.007 $\pm$ 0.000}}   & \multicolumn{1}{c|}{0.008 $\pm$ 0.000}  & 0.015 $\pm$ 0.001 & \multicolumn{1}{l|}{0.004 $\pm$ 0.005} & \multicolumn{1}{l|}{0.006 $\pm$ 0.010} & \multicolumn{1}{l|}{\textbf{0.003 $\pm$ 0.003}}   & \multicolumn{1}{l|}{0.003 $\pm$ 0.004}  & 0.005 $\pm$ 0.005                   \\
HalfCheetah           & \multicolumn{1}{c|}{\textbf{0.014 $\pm$ 0.001}} & \multicolumn{1}{c|}{0.019 $\pm$ 0.01}  & \multicolumn{1}{c|}{\textbf{0.014 $\pm$ 0.001}}   & \multicolumn{1}{c|}{0.015 $\pm$ 0.001}  & 0.022 $\pm$ 0.001 & \multicolumn{1}{l|}{0.005 $\pm$ 0.006} & \multicolumn{1}{l|}{0.007 $\pm$ 0.013} & \multicolumn{1}{l|}{\textbf{0.003 $\pm$ 0.004}}   & \multicolumn{1}{l|}{0.003 $\pm$ 0.005}  & 0.005 $\pm$ 0.008                   \\
Hopper                & \multicolumn{1}{c|}{\textbf{0.012 $\pm$ 0.001}} & \multicolumn{1}{c|}{0.017 $\pm$ 0.001} & \multicolumn{1}{c|}{\textbf{0.012 $\pm$ 0.001}}   & \multicolumn{1}{c|}{\textbf{0.012 $\pm$ 0.001}}  & 0.020 $\pm$ 0.001 & \multicolumn{1}{l|}{0.005 $\pm$ 0.006} & \multicolumn{1}{l|}{0.008 $\pm$ 0.012} & \multicolumn{1}{l|}{\textbf{0.003 $\pm$ 0.004}}   & \multicolumn{1}{l|}{\textbf{0.003 $\pm$ 0.004}}  & 0.005 $\pm$ 0.007                   \\
Kitchen               & \multicolumn{1}{c|}{\textbf{0.008 $\pm$ 0.000}} & \multicolumn{1}{c|}{0.014 $\pm$ 0.000} & \multicolumn{1}{c|}{0.009 $\pm$ 0.000}   & \multicolumn{1}{c|}{0.009$\pm$ 0.000}   & 0.017 $\pm$ 0.001 & \multicolumn{1}{l|}{0.006 $\pm$ 0.003} & \multicolumn{1}{l|}{0.009 $\pm$ 0.006} & \multicolumn{1}{l|}{\textbf{0.005 $\pm$ 0.003}}   & \multicolumn{1}{l|}{\textbf{0.005 $\pm$ 0.003}}  & 0.006 $\pm$ 0.004                   \\
Maze2d                & \multicolumn{1}{c|}{\textbf{0.008 $\pm$ 0.000}} & \multicolumn{1}{c|}{0.014 $\pm$ 0.000} & \multicolumn{1}{c|}{\textbf{0.008 $\pm$ 0.000}}   & \multicolumn{1}{c|}{\textbf{0.008 $\pm$ 0.000}}  & 0.015 $\pm$ 0.001 & \multicolumn{1}{l|}{\textbf{0.003 $\pm$ 0.003}} & \multicolumn{1}{l|}{0.006 $\pm$ 0.006} & \multicolumn{1}{l|}{\textbf{0.003 $\pm$ 0.003}}   & \multicolumn{1}{l|}{\textbf{0.003 $\pm$ 0.003}}  & 0.004 $\pm$ 0.004                   \\
Walker2d              & \multicolumn{1}{c|}{\textbf{0.013 $\pm$ 0.001}} & \multicolumn{1}{c|}{0.020 $\pm$ 0.001} & \multicolumn{1}{c|}{0.014 $\pm$ 0.001}   & \multicolumn{1}{c|}{0.014 $\pm$ 0.001}  & 0.022 $\pm$ 0.001 & \multicolumn{1}{l|}{0.005 $\pm$ 0.007} & \multicolumn{1}{l|}{0.006 $\pm$ 0.012} & \multicolumn{1}{l|}{\textbf{0.003 $\pm$ 0.004}}   & \multicolumn{1}{l|}{0.003 $\pm$ 0.005}  & 0.005 $\pm$ 0.008                  
\end{tabular}
}
\end{table*}

\section{Conclusion}

In summary, we introduced the DM and the HDM models, evolutions of the existing state-of-the-art sequence models, DT and HDT, respectively, by leveraging the potent Mamba architecture. 
We show that the DM does not rely on the sequence of RTG, or a reward function to perform.
Through our evaluation across seven diverse tasks within the D4RL benchmark and varying the demonstration data set, we demonstrate the superiority of these Mamba models over their Transformer-based predecessors in the majority of cases. Specifically, DM and HDM outperform their transformer counterparts in most settings, while also being faster to train and perform inference. Lastly, DM outperforms the baselines in most tasks, while being task-independent by not requiring user specified values. 
This advancement underscores the viability of Mamba architectures in behaviour cloning sequence modelling,
and opens the way to keep solving more complex sequence modelling problems through imitation learning.

\section*{Acknowledgments}

This work is supported by NOVA LINCS ref. UIDB/04516/2020\\ (https://doi.org/10.54499/UIDB/04516/2020) and ref. UIDP/04516/2020 (https://doi.org/10.54499/UIDP/04516/2020) with the financial support of FCT.IP, and also through the research grant 2022.14197.BD.

\end{document}